\documentclass{article}
\usepackage[nonatbib, final]{neurips18}
\usepackage{biblatex} 
\usepackage{graphicx}
\usepackage[utf8]{inputenc} 
\usepackage[T1]{fontenc}    
\usepackage{hyperref}       
\usepackage{url}            
\usepackage{booktabs}       
\usepackage{amsfonts}       
\usepackage{nicefrac}       
\usepackage{microtype}      
\pdfoutput=1

\title{Modularization of End-to-End Learning:\\Case Study in Arcade Games}

\author{
  Andrew Melnik\\
  CITEC\\
  Bielefeld University\\
  Bielefeld, Germany\\
  \texttt{andrew.melnik.papers@gmail.com} \\
  \And
  Sascha Fleer \\
  CITEC\\
  Bielefeld University\\
  Bielefeld, Germany\\
  \texttt{sfleer@techfak.uni-bielefeld.de} \\
  \AND
  Malte Schilling \\
  CITEC\\
  Bielefeld University\\
  Bielefeld, Germany\\
  \texttt{mschilli@techfak.uni-bielefeld.de} \\
  \And
  Helge Ritter \\
  CITEC\\
  Bielefeld University\\
  Bielefeld, Germany\\
  \texttt{helge@techfak.uni-bielefeld.de} \\
}

\begin{document}

\maketitle

\begin{abstract}
Complex environments and tasks pose a difficult problem for holistic \textit{end-to-end} learning approaches. Decomposition of an environment into interacting \textit{controllable} and \textit{non-controllable} objects allows supervised learning for \textit{non-controllable} objects and \textit{universal} value function approximator learning for \textit{controllable} objects. Such decomposition should lead to a shorter learning time and better generalisation capability. Here, we consider arcade-game environments as sets of interacting objects (\textit{controllable, non-controllable}) and propose a set of functional modules that are specialized on mastering different types of interactions in a broad range of environments. The modules utilize regression, supervised learning, and reinforcement learning algorithms. Results of this case study in different Atari games suggest that human-level performance can be achieved by a learning agent within a human amount of game experience (10-15 minutes game time) when a proper decomposition of an environment or a task is provided. However, automatization of such decomposition remains a challenging problem. This case study shows how a model of a causal structure underlying an environment or a task can benefit learning time and generalization capability of the agent, and argues in favor of exploiting modular structure in contrast to using pure \textit{end-to-end} learning approaches.\\

\textbf{Keywords:} Arcade Games; Causal Learning; Hierarchical Reinforcement Learning; End-to-End Learning.
\end{abstract}

\section{Introduction}

The priors that we put into the machine and the techniques we use for optimization are essential ingredients of learning. People can learn a new concept from a few examples, yet machine learning algorithms typically require thousands of examples to perform similarly \cite{lake2014towards, mccandlish2018empirical}. The compound nature of arcade-game environments leads to an exponentially increasing learning time with the increasing complexity of an environment or a task for \textit{end-to-end} learning approaches. \textit{End-to-end} learning without priors requires a large number of interactive steps to attain human-level performance even in seemingly straightforward game environments \cite{mnih2015human}.

\newpage
Biological agents do not learn by a single, global optimization principle within a uniform neural network. Rather, their brains are modular, with distinct but interacting subsystems working in parallel and contributing to an overall emergent behavior \cite{hassabis2017neuroscience, anderson2010neural, schilling2018approach}. In our framework, we use a simplified representation of such functional subsystems and call them \textit{functional modules}. Each functional module allows an artificial agent to learn a specific type of interaction primitive in an environment. We assume, that there is a moderate number of such interaction primitives required to fully describe interactions in any physical environment. Therefore, only a moderate number of functional modules is required for optimization of agent's behaviour in an environment. 

In the current study, we describe a set of functional modules for specific types of interaction primitives, which are common to a broad range of arcade environments. Our goal is to come up with a reusable toolset of functional modules which can adapt their predefined functionality to a specific environment using a minimal number of observation samples. This will allow shifting the learning problem to a higher level of disentangling rewarding interactions in an environment.

\section{Framework}
\begin{figure}
\centering
\includegraphics[width=0.87\textwidth]{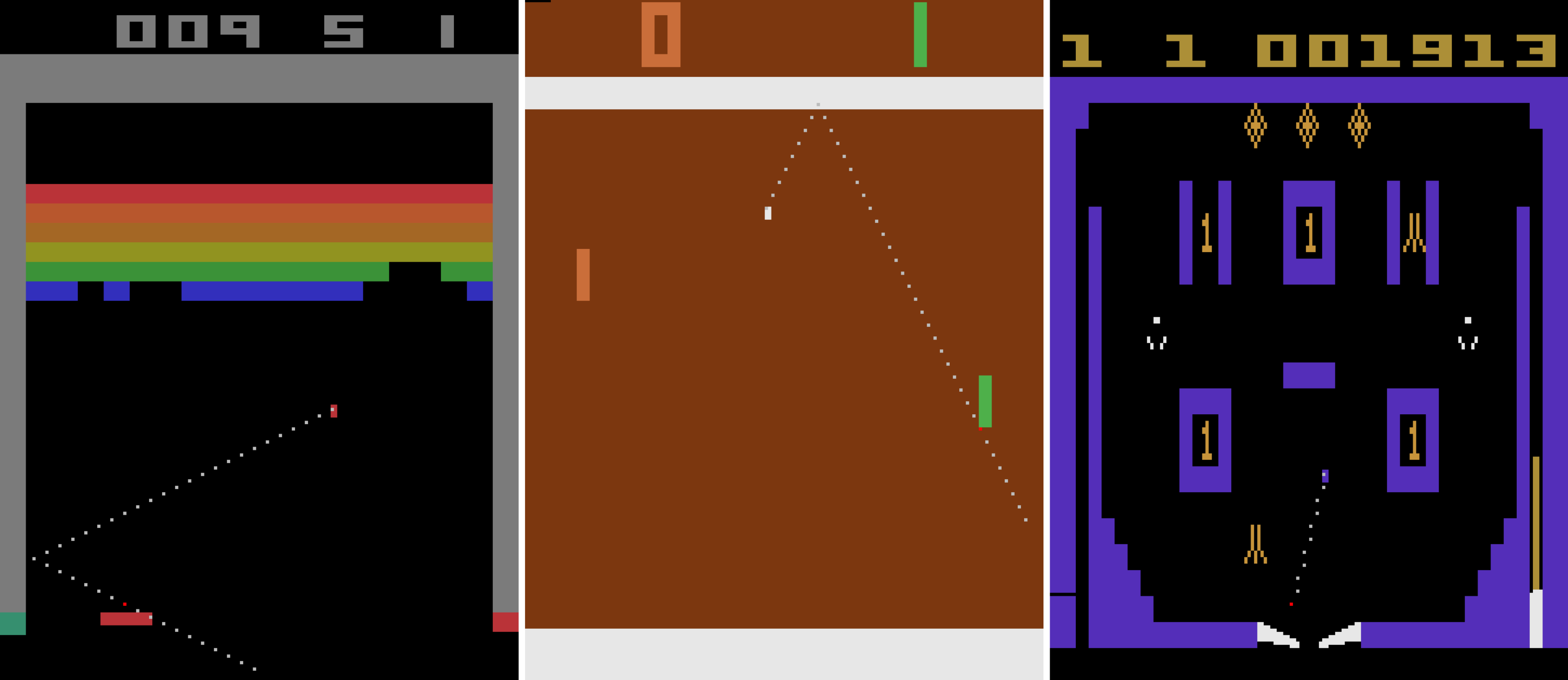}
\caption{We tested our framework on the OpenAI-Gym-Atari environments: Breakout, Pong, and Pinball. The trained framework predicts a future trajectory of a ball (white dotted lines) and controls a paddle to bring these objects to a hit superposition that maximizes an expected reward. Video: \href{https://rebrand.ly/ataridemo}{\underline{https://rebrand.ly/ataridemo}}}
\label{fig:env}
\end{figure}

\begin{figure}
\centering
\includegraphics[width=0.9\textwidth]{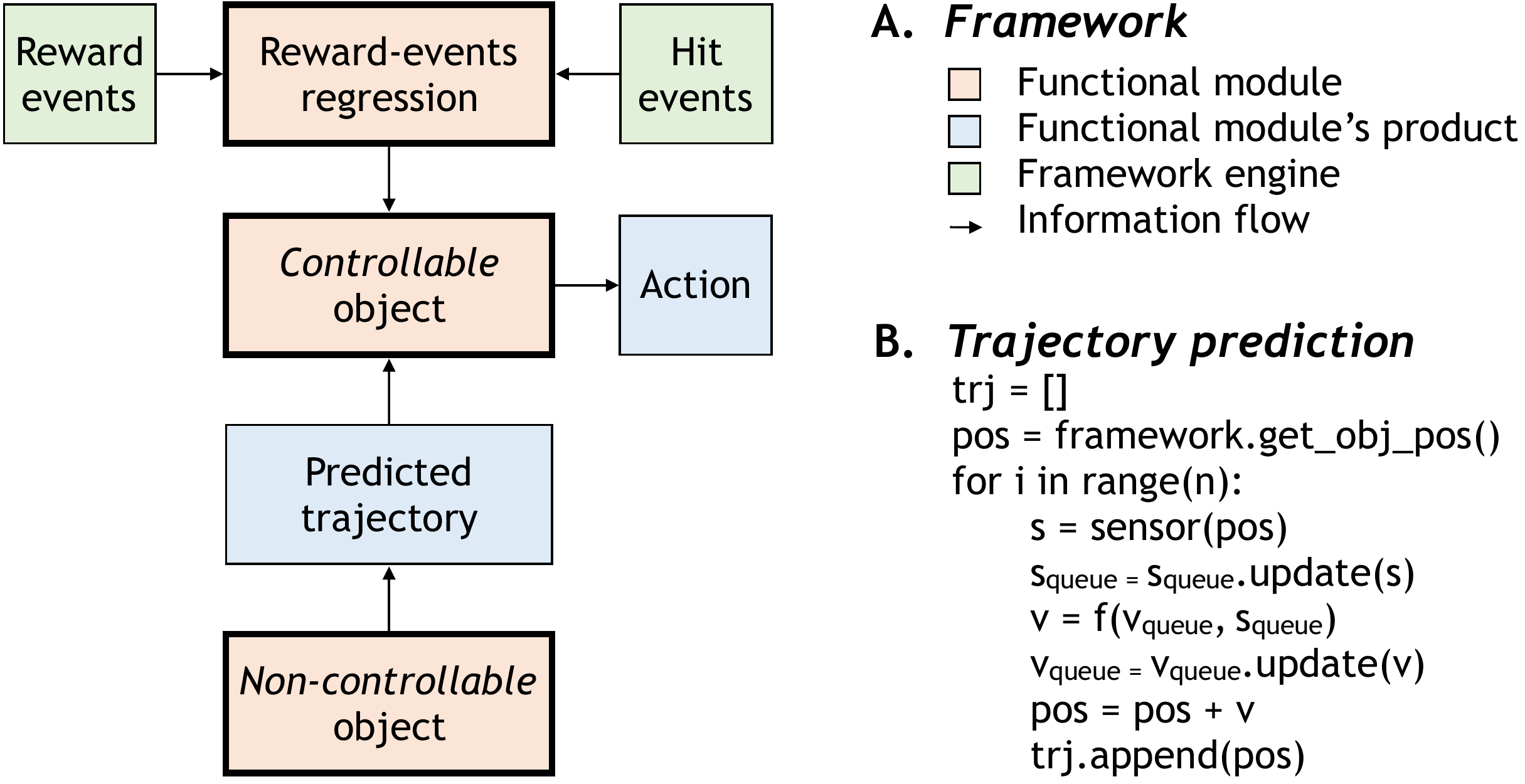}
\caption{\textbf{A.} Overview. \textbf{B.} Trajectory-prediction pseudo code of the non-controllable-object module. \textit{n} - number of recursive prediction steps; \textit{pos} - anterior position of a predicted trajectory at the prediction step \textit{i}; \textit{s} - observation from the sensor at the step \textit{i}; \textit{v} - predicted velocity vector at the step \textit{i}.}
\label{fig:arch}
\end{figure}

We tested our framework on three OpenAI-Gym-Atari \cite{brockman2016openai} environments: Breakout, Pong, and Pinball (Fig.~\ref{fig:env}). While these game-environments share some generic interaction primitives which can be learned by the general-purpose functional modules we proposed (Fig.~\ref{fig:arch}), they still are very different in appearance and in-game dynamics.

The framework uses object based representation. Each object (e.g. ball, paddle, wall, etc.) has a data structure containing information about the shape of the object, object type, and previous positions and velocities. Additionally the framework stores the area in which the object was observed throughout the gameplay and set of available actions for \textit{controllable} objects. 

The exact techniques for extracting object-related information from the pixel-based input are not in the focus of the current study. Therefore, we do not provide a general approach for extraction of objects and their features from a pixel representation of arcade-game environments. For the given environments (Breakout, Pong, Video Pinball) we apply adapted filters, like color selection, difference between sequential frames, and shape matches to extract object-related information from screens. Such information can also be provided by a simulation engine, as seen in other studies \cite{kansky2017schema, kidzinski2018learning}.

Interaction primitives are processed by a set of functional modules (Fig.~\ref{fig:arch}), which were the same for all three game-environments in the study. Input-output relationships in the modules are formed by regression, supervised learning, or reinforcement learning algorithms.

\newpage
\subsection{Module 1: Non-controllable object}

One important and general property of \textit{non-controllable} objects in arcade-game environments is their movement trajectories. It is necessary, for example, to predictively chase an object, avoid a collision, etc. Therefore, we designed a functional module which predicts the future movement trajectory of an object. In each game-environment, the functional module uses supervised learning to train a neural network on past trajectories to predict a future trajectory of the object by rolling out a sequence of \textit{n} recursive prediction steps (Fig.~\ref{fig:arch}B). The network has two hidden layers followed by a split into two streams, one for \textit{Vx} and one for \textit{Vy} velocity output dimensions, with one hidden layer each. All hidden layers use the ReLu activation function.

We added a \textit{sensor} to predict outcomes of interactions with other objects (e.g. collisions) at the trajectory path of the \textit{non-controllable object}. The \textit{sensor} detects overlapping of \textit{m} neighboring pixels with other objects (0 - no overlapping; 1 - overlapping detected) at each recursive prediction steps. The information from the sensor is represented as a vector of \textit{m} binary values. Therefore, the module learns to predict a trajectory while taking into account interaction outcomes with other objects.

The neural network input is a vector containing a queue of \textit{k} previous velocity vectors and sensor observations concatenated and flattened in a 1-D vector. At each recursive step the neural network outputs a predicted velocity vector of two values \textit{(Vx, Vy)}. And the framework updates the velocity queue, updates the anterior position point of the trajectory, appends it to the stack of trajectory positions of the \textit{non-controllable object}, and updates the sensor queue (Fig.~\ref{fig:arch}B). 

\subsection{Module 2: Reward-events regression}

Shape is an important property of an object that influences interaction with other objects. In the framework, objects have a bitmap representation of their shapes. The framework produces an event for overlapping of \textit{controllable} and \textit{non-controllable} objects' shapes (e.g., ball-paddle hit event). The \textit{reward-events regression} module (Fig.~\ref{fig:arch}) collects statistics on contact points with the \textit{controllable} object and delayed rewards, and builds the reward regression for contact points, relative to the center of the \textit{controllable object's} shape. Points of the shape associated with higher reward values are selected with a higher probability as the intersection point for the predicted trajectory of an \textit{non-controllable object}. 

For example, in the Pong game, a contact point at the middle of the paddle (with a ball) is associated with the zero expected reward value according to the collected statistics, as reflecting the ball at the middle of the paddle never leads to score points for any of the players. A potential hit event farther than the edge of the paddle has a negative expected reward value, as it leads to missing a ball. A hit event at the edge of the paddle has the highest expected reward value, as reflection of the ball at the edge of the paddle leads to a steep attack angle and a high probability of scoring. 

\subsection{Module 3: Controllable object}

The \textit{controllable-object} module (Fig.~\ref{fig:arch}A) operates a directly-controllable object in a game environment (e.g. a paddle) by specifying the goal position for the \textit{controllable} object in the environment \cite{kulkarni2016hierarchical}. When a predicted trajectory of a \textit{non-controllable} object (e.g. a ball) intersects the area accessible by the \textit{controllable} object, the module utilizes the intersection position as the goal position for the \textit{controllable} object. The center of the \textit{controllable} object's shape is the current position of the \textit{controllable} object. The \textit{reward-events regression} module specifies the point on the \textit{controllable} object's shape that should intersect with the goal position. The module also utilizes \textit{k} previous actions applied to the \textit{controllable} object to take into account the dynamics of the object. In every time step, the module outputs an action for the \textit{controllable} object, leading to approaching the goal according to a learned universal-value policy \cite{schaul2015universal}. The module uses Hindsight Experience Replay \cite{andrychowicz2017hindsight} to minimize a number of experience samples required for convergence. 

\section{Performance}

\begin{figure}
\centering
\includegraphics[width=1.0\textwidth]{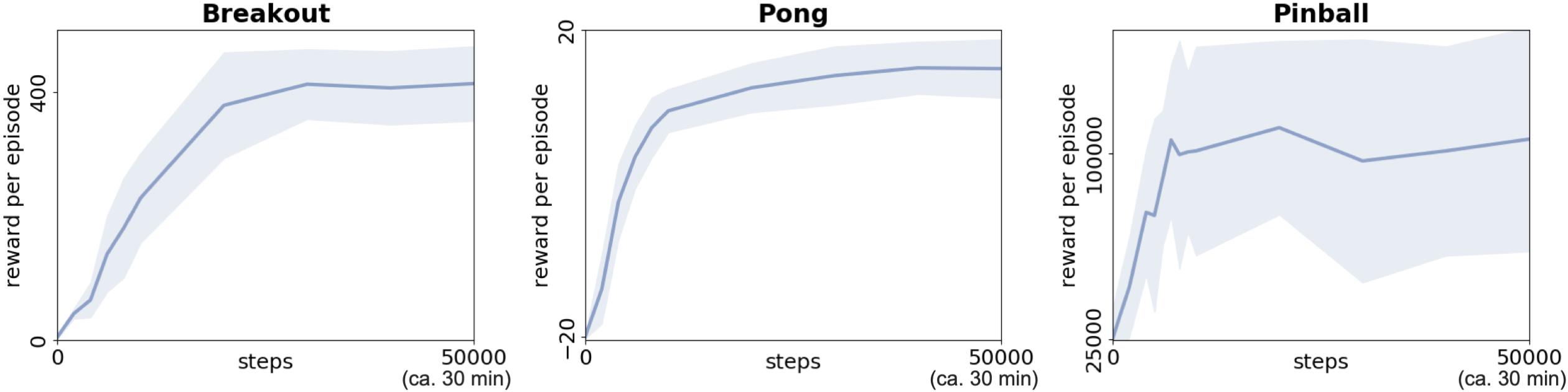}
\caption{Learning curves of the framework in Atari-arcade-game environments.}
\label{fig:plots}
\end{figure}

Our framework achieves human-level control in less than 25 000 steps (frame skip = 2) (Fig.~\ref{fig:plots}) in such Arcade Games as Breakout, Pong, and Pinball. This is approximately equal to 10-15 min of real-time human play (60 Hz), which is close to how fast human players are mastering these games. In contrast, classic end-to-end reinforcement learning models with pixel- or object-based representations without modularization require about 25 million steps \cite{hessel2017rainbow} or more to achieve similar performance.

\section{Discussion}

In the current case study, we proposed several general-purpose functional modules applicable to a broad range of arcade-game environments. The modules can adapt their predefined functionality to a specific environment using a minimal number of observation samples. We demonstrated that the framework is capable of acquiring human-level performance within a human amount of game experience (10-15 minutes game time).

A number of possible interaction primitives of physical objects is limited and moderate. Extending the framework with additional general-purpose functional modules, covering a broader range of interaction primitives, will allow for human-level control in a broader range of arcade environments. Following up on this line of research will allow to shift the scaling challenges of reinforcement learning to the question of how the number of to-be-covered interaction primitives of physical objects compares to the complexity of the environment. Most likely, there are different ways of defining such interaction primitives, serving different trade-offs between, e.g., minimal number, generalization capability and speed of learning. To gain deeper insights into the structure of these trade-offs would be an ambitious longer-term goal, both for a comprehensive theory of reinforcement learning and for a principled tailoring of practical learning architectures \cite{melnik2018world}.

We assume a taxonomy of complexity of interaction primitives required for mastering arcade-game environments. This means that some arcade-game environments may require more functional modules than others for a successful mastering with the framework. The influential study by Mnih et al. \cite{mnih2015human} demonstrated the performance of pixel based DQN with respect to a professional human games tester, where, “Video Pinball” and “Breakout” environments were mastered by the DQN approach at the considerably higher level than the professional human games tester (>1000\ \%); “Pong” near the human level (132\ \%); and “Montezuma's Revenge” at the random play level (0\ \%). Their comparison provides a first approximation for such a taxonomy of complexity of games.

In the study, we converted pixel-based representation of the environments into object-based representation. Such composable object-based representations and a neural network architecture allow factorizing object dynamics into pairwise interactions, as seen in other learning approaches \cite{chang2016compositional, hamrick2017metacontrol}. We also expect good generalization properties of the framework for subtle changes in the environment, such as paddle offset, middle wall, etc. \cite{kansky2017schema}.

This work proposes a case study where Atari games are modularized into parts (\textit{controllable, non-controllable} objects). Such a modular approach may allow shifting the learning problem to a higher level of disentangling rewarding interactions in an environment. This case study shows how a model of a causal structure underlying an environment or a task can benefit learning time and generalization capability of the agent, and argues in favor of exploiting modular structure in contrast to using pure \textit{end-to-end} learning approaches.

\section*{Acknowledgments}
We acknowledge the support by German Research Council (DFG), grant EXC 277. 

\AtNextBibliography{\small}
\printbibliography
\end{document}